# Formal Model of Uncertainty for Possibilistic Rules

Arthur Ramer
University of Oklahoma, Norman, OK 73019

Leslie Lander
SUNY-Binghamton, Binghamton, NY 13902-6000

## OVERVIEW

Given a universe of discourse $X$—a domain of possible outcomes—an experiment may consist of selecting one of its elements, subject to the operation of chance, or of observing the elements, subject to imprecision.

A priori uncertainty about the actual result of the experiment may be quantified, representing either the likelihood of the choice of $x \in X$ or the degree to which any such $x \in X$ would be suitable as a description of the outcome. The former case corresponds to a probability distribution, while the latter gives a possibility assignment on $X$.

The study of such assignments and their properties falls within the purview of possibility theory [DP88, Y80, Z78]. It, like probability theory, assigns values between 0 and 1 to express likelihoods of outcomes. Here, however, the similarity ends. Possibility theory uses the maximum and minimum functions to combine uncertainties, whereas probability theory uses the plus and times operations. This leads to very dissimilar theories in terms of analytical framework, even though they share several semantic concepts. One of the shared concepts consists of expressing quantitatively the uncertainty associated with a given distribution. In probability theory its value corresponds to the gain of information that would result from conducting an experiment and ascertaining an actual result. This gain of information can equally well be viewed as a decrease in uncertainty about the outcome of an experiment. In this case the standard measure of information, and thus uncertainty, is Shannon entropy [AD75, G77]. It enjoys several advantages—it is characterized uniquely by a few, very natural properties, and it can be conveniently used in decision processes. This application is based on the principle of maximum entropy; it has become a popular method of relating decisions to uncertainty.

This paper demonstrates that an equally integrated theory can be built on the foundation of possibility theory. We first show how to define measures of information and uncertainty for possibility assignments. Next we construct an information-based metric on the space of all possibility distributions defined on a given domain. It allows us to capture the notion of proximity in information content among the distributions. Lastly, we show that all the above constructions can be carried out for 'continuous distributions'—possibility assignments on arbitrary measurable domains. We consider this step very significant—finite domains of discourse are but approximations of the real-life infinite domains. If possibility theory is to represent real world situations, it must handle continuous distributions both directly and through finite approximations.

In the last section we discuss a principle of maximum uncertainty for possibility distributions. We show how such a principle could be formalized as an inference rule. We also suggest it could be derived as a consequence of simple assumptions about combining information.

We would like to mention that possibility assignments can be viewed as fuzzy sets and that every fuzzy set gives rise to an assignment of possibilities. This correspondence has far reaching consequences in logic and in control theory. Our treatment here is independent of any special interpretation; in particular we speak of possibility distributions and possibility measures, defining them as measurable mappings into the interval $[0, 1]$.

Our presentation is intended as a self-contained, albeit terse summary. Topics discussed were selected with care, to demonstrate both the completeness and a certain elegance of the theory. Proofs are not included; we only offer illustrative examples.

## 1 POSSIBILITY DISTRIBUTIONS AND MEASURES

### 1.1 DISCRETE DOMAINS

We use the model of possibility theory introduced in [Z78]. The domain of discourse can be any finite or finitely measurable set. Here we discuss a finite domain $X$ and define a *possibility distribution* as a function $\pi : X \to [0,1]$ such that $\max_{x \in X} \pi(x) = 1$. It expresses an assignment of possibility values $\pi(x)$ to elementary events $x \in X$. We extend it to arbitrary subsets $Y \subset X$ putting $\pi(Y) = \max_{x \in Y} \pi(x)$.[1]

Given two domains $X$ and $Y$ and two independent possibility assignments $\pi_1 : X \to [0,1]$, $\pi_2 : Y \to [0,1]$ we define a joint distribution
$$\pi_1 \otimes \pi_2 : (x,y) \mapsto \min(\pi_1(x), \pi_2(y)).$$
Given an arbitrary assignment $\pi$ on a product space $X \times Y$ we define its marginal assignments $\pi'$ on $X$ and $\pi''$ on $Y$ as
$$\pi'(x) = \max_{y \in Y} \pi(x,y), \qquad \pi''(y) = \max_{x \in X} \pi(x,y).$$
It is also convenient to define an *extension* of $\pi$ from its domain $X$ to a larger set $Y \supset X$. We put $\pi^Y(y) = \pi(y)$, $y \in X$ and $\pi^Y(y) = 0$ otherwise. Lastly, given $X = \{x_1, \ldots, x_n\}$ and a permutation $s$ of $\{1, \ldots, n\}$, we define a possibility assignment $s(\pi)$
$$s(\pi)(x_i) = \pi(x_{s(i)}).$$

### 1.2 CONTINUOUS DOMAINS

This structure generalizes to an arbitrary $X$ endowed with a finite (Lebesgue) measure. The assignment becomes a measurable function $f : X \to [0,1]$, defining the possibility of $Y \subset X$ as $\sup_{x \in Y} f(x)$. By analogy with probability theory, we term such structures *continuous* possibility assignments or distributions. Joint, marginal and extended distributions are now defined using *sup* and *inf* instead of *max* and *min*. Lastly, we generalize permutations of $X$ to measure-preserving transformations $s$, putting $s(f)(x) = f(s(x))$. A transformation corresponding to sorting discrete values is of particular interest. For $f$ defined on $X$ we want $\tilde{f}$ to be a descending equivalent of $f$, defined on a real interval of the same measure as $X$. For definiteness, we can make the origin the left end-point of the interval and have $\tilde{f}$ decrease monotonically. Measure-preserving implies that $\tilde{f}$ 'stays' above any given value $\alpha$, $0 \leq \alpha \leq 1$, over the same space as the original function [2] and leads to a classical construction [HLP34].

We put $P(y) = \mathcal{M}\{x : f(x) \geq y\}$, where $\mathcal{M}$ is a standard measure on $[0,1]$ and define $\tilde{f}(x) = P^{-1}(x)$. As an illustration let us consider two examples.

**Example**
$$f(x) = \begin{cases} 2x, & 0 \leq x \leq 0.5, \\ 2 - 2x, & \text{otherwise.} \end{cases}$$

---
[1] The fuzzy interpretation is obtained by treating the pair $(X, \pi)$ as a fuzzy subset of $X$ and $\{x : \pi(x) \geq \alpha\}$ as its $\alpha$-cuts.

[2] All $\alpha$-cuts [DP88] of $f$ are of the same size (have the same measure) as $\alpha$-cuts of $\tilde{f}$.

Here $P(y) = 1 - y$ and $\tilde{f}(x) = 1 - x$. It is immediate that $\tilde{f}(x) \geq \alpha$ over the set of the same measure as the set where $f(x) \geq \alpha$.

**Example** $f(x) = 4(x - \frac{1}{2})^2 = 4x^2 - 4x + 1$.
Now $P(y)$ represents the *combined* length of the intervals where $f(x)$ is $\geq y$.

Since $\{x : f(x) \geq y\} = [0, \frac{1-\sqrt{y}}{2}] \cup [\frac{1+\sqrt{y}}{2}, 1]$, we have $P(y) = 1 - \sqrt{y}$. Therefore $y = (1 - P(y))^2$ and $\tilde{f}(x) = (1-x)^2$.

## 2 INFORMATION FUNCTIONS IN POSSIBILITY THEORY

### 2.1 UNCERTAINTY

The structure outlined above provides the *possibilistic* context for the quantification of the notions of uncertainty and information. We view the mapping $\pi$ as assigning a degree of assurance or certainty that an element of $X$ is the outcome of an experiment. That experiment would consist of selecting $x \in X$ as a representative (perhaps unique) object of discourse. A priori we know only the distribution $\pi$; to determine $x \in X$ means to remove uncertainty about the result, thus entailing a gain of information. We would be particularly interested in quantifying that gain of information, averaged over the complete distribution $\pi$. That would also express the overall value of uncertainty inherent in the complete distribution $\pi$. Accordingly, we intend to define an information function $I$ which assigns a nonnegative real value to an arbitrary distribution $\pi$. Following established principles of information theory, [AD75, G77] we stipulate that such an information function satisfies certain standard properties. Specifically, we require

| | | |
|---|---|---|
| *additivity* | $I(\pi_1 \otimes \pi_2) =$ | $I(\pi_1) + I(\pi_2)$ |
| *subadditivity* | $I(\pi) \leq$ | $I(\pi') + I(\pi'')$ |
| *symmetry* | $I(s(\pi)) =$ | $I(\pi)$ |
| *expansibility* | $I(\pi^Y) =$ | $I(\pi)$ |

It turns out that these properties essentially characterize the admissible information functions [KM87,RL87]. Here we discuss the discrete case of $X = \{x_1, \ldots, x_n\}$. Let $\tilde{p}_1 \geq \tilde{p}_2 \geq \ldots \geq \tilde{p}_n$ be a descending sequence formed from the values $\pi(x_1), \ldots, \pi(x_n)$. Then, up to a multiplicative constant:

**Theorem** All information functions on $X$ are of the form
$$\begin{aligned} I(\pi) &= \sum_{i=1}^{n-1} (\tau(\tilde{p}_i) - \tau(\tilde{p}_{i+1})) \log i \\ &= \sum_{i=2}^{n} \tau(\tilde{p}_i) \nabla \log i \end{aligned}$$
where $\tau$ is a nondecreasing mapping of $[0,1]$ onto itself. $I(\pi)$ is continuous (as a functional on the space of distributions) iff $\tau$ is a continuous deformation of $[0,1]$.

The formula can be derived from functional equations representing the properties of information. In particular, the presence of $\log i$ comes from additivity, while the differences $\tau(\tilde{p}_i) - \tau(\tilde{p}_{i+1})$ reflect the use of the *max* and *min* operations.

By analogy with Shannon theory we may also impose a linear interpolation property on $I(\pi)$ [KM87]. We then obtain a particularly simple expression, named *U*-uncertainty [HK82]

$$U(\pi) = \sum (\tilde{p}_i - \tilde{p}_{i+1}) \log i = \sum \tilde{p}_i \nabla \log i.$$

It follows by taking $\tau$ to be the identity mapping and we shall continue to do that in the remainder of the paper; however, all the results can be extended to an arbitrary $\tau$.

We observe that the distribution which carries the highest uncertainty value consists of assigning possibility 1 to all the events in $X$. It states that, a priori, every event is fully possible. This distribution, carrying no prior information, can be considered the *most uninformed* one.

## 2.2 INFORMATION DISTANCE

U-uncertainty serves to define various information distances [HK83, R90] between two distributions $\pi$ and $\rho$ defined on the same domain $X$. If $\pi(x) \leq \rho(x)$, we put

$$g(\pi, \rho) = U(\rho) - U(\pi).$$

For the general case, given $\pi$ and $\rho$, we first define their lattice meet and join

$$\pi \wedge \rho : x \mapsto \min(\pi(x), \rho(x)),$$
$$\pi \vee \rho : x \mapsto \max(\pi(x), \rho(x)).$$

We then put

$$G(\pi, \rho) = g(\pi, \pi \vee \rho) + g(\rho, \pi \vee \rho),$$
$$H(\pi, \rho) = g(\pi \wedge \rho, \pi) + g(\pi \wedge \rho, \rho),$$
$$K(\pi, \rho) = \max(g(\pi, \pi \vee \rho), g(\rho, \pi \vee \rho)).$$

These functions have several attractive properties:

**Theorem** Both $G$ and $K$ define metric distances on the space of all possibility distributions (on a given domain). $H$ is additive in both arguments

$$H(\pi_1 \otimes \pi_2, \rho_1 \otimes \rho_2) = H(\pi_1, \rho_1) + H(\pi_2, \rho_2).$$

## 3  DESIGN OF CONTINUOUS POSSIBILITY INFORMATION

We shall now extend the previous definitions to arbitrary measurable domains. To avoid technical complications, we consider only the special, albeit typical case where $X$ is the unit interval. Now a possibility distribution is a function $f : [0,1] \to [0,1]$ such that $\sup_{x \in [0,1]} f(x) = 1$. Although in a variety of practical situations it is sufficient to consider only continuous functions, we do not make that restriction.

As a first step the discrete formula $U(\pi) = \sum p_i \nabla \log i$ suggests forming an expression like $\int_0^1 \tilde{f}(x) d\ln x$, where $\tilde{f}$ is a suitable 'decreasing sorted' equivalent of $f$, while $d\ln x$ substitutes $\nabla \log x$. The latter quantity simply represents $x^{-1} dx$, while for $\tilde{f}$ we use a descending rearrangement of $f$.

Using this definition we can consider $\int_0^1 \frac{\tilde{f}(x)}{x} dx$ as a candidate expression for the value of information. Unfortunately, $\tilde{f}(x)$ is equal to 1 at 0, and the integral above diverges. A solution can be found through a technique that has been used in probability theory [G77], which is to use the information distance between a given density and the uniform one. In possibility theory we consider a constant function $f(x) \equiv 1$ as representing a uniform distribution. It is also the most 'uninformed' one—its discrete form clearly attains maximum U-uncertainty. Our final formula becomes

$$I(f) = \int_0^1 \frac{1 - \tilde{f}(x)}{x} dx.$$

This integral is well defined and avoids the annoying singularity at 0. We demonstrate its use on a class of polynomial functions.

**Example**  Let us consider possibility distributions represented by $f(x) = x^n$, $n = 0, 1, \ldots$. Writing $J_n = I(x^n)$ and remembering that $\tilde{x^n} = (1-x)^n$, let us first compute $J_n - J_{n-1}$

$$\int_0^1 \frac{(1 - (1-x)^n) - (1 - (1-x)^{n-1})}{x} dx =$$

$$\int_0^1 \frac{(1-x)^{n-1} - (1-x)^n}{x} dx = \int_0^1 (1-x)^{n-1} dx = \frac{1}{n}$$

As $J_0 = 0$ we find that $J_n = 1 + \frac{1}{2} + \cdots + \frac{1}{n} = H_n$, the $n^{\text{th}}$ harmonic number.

## 4  PROPERTIES OF CONTINUOUS INFORMATION MEASURES

We summarize the properties of $I(f)$ in the next two theorems; $f, f_1, \ldots$ stand for continuous distributions, $f_1 \otimes f_2$ for their *min*-product and $f'$ and $f''$ for the projections of $f$ when it is defined on a product space.

**Theorem** $I(f)$ is

| | | |
|---|---|---|
| *additive* | $I(f_1 \otimes f_2) =$ | $I(f_1) + I(f_2)$ |
| *superadditive* | $I(f) \leq$ | $I(f') + I(f'')$ |
| *symmetric* | $I(s(f)) =$ | $I(f)$ |
| *expansible* | $I(f^Y) =$ | $I(f)$ |

Superadditivity of $I$ (replacing subadditivity of $U$) is due to the minus sign in the formula that defines it.

Using $I(f)$ we can define continuous extensions of the information distances $g$, $G$, $H$ and $K$. As in the discrete case, $G$ and $K$ are metric distances, while $H$ is additive in both arguments. We shall demonstrate additivity of $I$ with an example.

**Example** We use as an example $f = g = x^\alpha$, $\alpha \geq 0$. We put $h(x,y) = \min(x^\alpha, y^\alpha)$ and find $\bar{h}(t) = (1 - \sqrt{t})^\alpha$. Then $I(h) = \int_0^1 \frac{1-(1-\sqrt{t})^\alpha}{t} dt$ which, after the substitution $u = \sqrt{t}$ becomes

$$\int_0^1 \frac{1-(1-u)^\alpha}{u^2} \cdot 2u\, du = 2\int_0^1 \frac{1-(1-u)^\alpha}{u} du = 2I(f).$$

**Theorem** $I(f)$ can be approximated as a limit of $U(p_n)$, where the $p_n$ are discrete distributions approximating $f$. [3]

This theorem confirms that we are justified using discrete possibility distributions in uncertainty computations. Their information values approximate consistently an idealized value of a putative continuous distribution. Already a non-trivial example is offered by a linear function.

**Example** We select $f(x) = 1 - x$ and approximate it using the values at $\frac{1}{n}, \frac{2}{n}, \ldots 1$. The approximating distributions are $\pi^{(n)} = (\frac{n-1}{n}, \frac{n-2}{n}, \ldots 1)$, thus

$$U(\pi^{(n)}) = \sum(p_i - p_{i+1})\ln i =$$

$$\sum(\frac{n-i+1}{n} - \frac{n-i}{n})\ln i = \frac{1}{n}\sum \ln i = \frac{1}{n}\ln n!$$

From Stirling's formula $U(\pi^{(n)}) \sim \ln n - 1$ and $I(\pi^{(n)}) = \ln n - U(\pi^{(n)}) = 1$ which agrees with $I(f)$.

## 5 PRINCIPLE OF MAXIMUM UNCERTAINTY

The decision rule forming the principle of maximum uncertainty can be stated independently of any specific theory used to capture the notions of randomness, vagueness or imprecision [G77, SJ80, J82]. We only need to assume that such randomness, vagueness or imprecision is expressed in the form of a numerical *information* function. The rule can be enhanced if, in addition, an information distance function is available.

The principle offers a method of selecting a distribution subject to certain constraints, usually presented as systems of linear equations on the parameters of an unknown distribution. Such constraints define a set of admissible distributions, and the choice from among those, the reasoning continues, should be made without introducing extraneous information, or should be as 'uninformed' as possible. Thus we should select a distribution of the maximum uncertainty value.

A variation of the rule occurs when we are given a 'prior' distribution and are required to replace it with a 'posterior' distribution, subject to admissibility criteria. Now we select the distribution for which the distance from the current one ('prior') reaches a *minimum*. The earlier case can be viewed as selecting a distribution closest to a hypothetical 'least informed' distribution.

In possibility theory such a principle would state that, given a prior assignment of possibility values and certain constraints on the posterior assignemnt, we should select the latter as the closest admissible assignment. The proximity here is expressed through the possibilistic information distance. If there is no known or assumed prior assignemnt, we should consider the distance from the most 'uninformed' possibility distribution, which is given by assigning a constant value 1 to every element of the domain of discourse. It clearly has the highest value of $U$-uncertainty; it also agrees with the intuitive perception that, in the absence of constraints, every choice should be accorded maximum possibility.

A similar method of determining distributions holds valid in probability theory [SJ80]. There it can be also shown that any reasonable selection based on maximization must be based on an information measure. Our current research aims to show that also in possibility theory decisions based on information measures stand privileged.

## REFERENCES

**AD75** ACZEL,J. and DAROCZY,Z., 1975, *On measures of information and their characterization*, Academic Press, New York.

**DP88** DUBOIS,D. and PRADE,H., 1988, *Possibility theory*, Plenum Press, New York.

**DP87** DUBOIS,D. and PRADE,H., 1987, Properties of measures of information in evidence and possibility theories, *Fuzzy Sets Syst.*, 24(2).

**G77** GUIASU,S., 1977, *Information Theory and Applications*, McGraw Hill, New York.

**HK83** HIGASHI,M. and KLIR,G., 1983, On the notion of distance representing information closeness, *Int. J. Gen. Syst.*, 9(1).

**HK82** HIGASHI,M. and KLIR,G., 1982, Measures of uncertainty and information based on possibility distributions *Int. J. Gen. Syst.*, 8(3).

**HLP34** HARDY,G., LITTLEWOOD,J. and POLYA,G., 1934, *Inequalities*,

---

[3] For sufficiently uniform approximations.

Cambridge University Press, Cambridge .

**J82**  JAYNES, E., 1982, On the rationale of maximum entropy methods, *Proc. IEEE*, 70.

**KM87**  KLIR, G. and MARIANO, M., 1987, On the uniqueness of possibilistic measure of uncertainty and information, *Fuzzy Sets Syst.*, 24(2).

**R90**  RAMER, A., 1990, Structure of possibilistic information metrics and distances, *Int. J. Gen. Syst.*, 17(1), 18(1).

**R89**  RAMER, A., 1989, Concepts of fuzzy information measures on continuous domains, *Int. J. Gen. Syst.*, 17(2-3).

**RL87**  RAMER, A. and LANDER, L., 1987, Classification of possibilistic uncertainty and information functions, *Fuzzy Sets Syst.*, 24(2).

**SJ80**  SHORE, J. and JOHNSON, R., 1980, Axiomatic derivation of the principle of maximum entropy and the principle of minimum cross-entropy, *IEEE Trans. Inf. Theory*, IT-26.

**Y80**  YAGER, Y., 1980, Aspects of possibilistic uncertainty, *Int. J. Man-Machine Studies*, 12.

**Z78**  ZADEH, L., 1978, Fuzzy sets as a basis for a theory of possibility, *Fuzzy Sets Syst.*, 3(1).